\documentclass[conference]{IEEEtran}
\IEEEoverridecommandlockouts
\usepackage[top=54pt, left=54pt, right=54pt, bottom=54pt]{geometry}
\usepackage{algorithmic}
\usepackage{enumitem}
\usepackage{cite}
\usepackage{amsmath,amssymb,amsfonts}
\usepackage[ruled,vlined]{algorithm2e}
\usepackage{graphicx}
\usepackage{textcomp}
\usepackage{xcolor}
\usepackage{multirow}
\usepackage{booktabs}
\usepackage{subcaption}
\usepackage{hyperref}
\usepackage{todonotes}
\usepackage{color,soul}
\usepackage{xcolor}
\usepackage{svg}
\usepackage{nicematrix}

\hypersetup{
     colorlinks = true,
     linkcolor = blue,
     anchorcolor = blue,
     citecolor = green,
     filecolor = magenta,
     urlcolor = cyan
     }
\def\BibTeX{{\rm B\kern-.05em{\sc i\kern-.025em b}\kern-.08em
    T\kern-.1667em\lower.7ex\hbox{E}\kern-.125emX}}

\newcommand*\rot{\rotatebox{90}}

\begin{document}

\title{ \vspace{0.4in} Radar-Lidar Fusion for Object Detection by Designing Effective Convolution Networks
\\
}
\author{\IEEEauthorblockN{Farzeen Munir\textsuperscript{1,2}, Shoaib Azam\textsuperscript{1,2}, Tomasz Kucner\textsuperscript{1}, Ville Kyrki\textsuperscript{1}, Moongu Jeon\textsuperscript{3}}
\IEEEauthorblockA{\textsuperscript{1}Department of Electrical Engineering and Automation,
Aalto University,
 Finland\\
}
\IEEEauthorblockA{\textsuperscript{2}Finnish Center for Artificial Intelligence,
Finland\\
}
\IEEEauthorblockA{\textsuperscript{3}School of Electrical Engineering and Computer Science,
Gwangju Institute of Science and Technology,
South Korea\\
Email: {(farzeen.munir, shoaib.azam, tomasz.kucner, ville.kyrki)@aalto.fi, mgjeon@gist.ac.kr}
}

}
% \author{\IEEEauthorblockN{Farzeen Munir}
% \and
% \and
% \and
% \and
% \and

\maketitle

\begin{abstract}
% Learning world models in terms of perception of the environment is instrumental in developing safe autonomous vehicles. 
% Object detection, a fundamental building block of perception, provides ego vehicle information about the surroundings to plan a safe route. Although cameras and Lidar have contributed to building a propitious perception system, yet have restricted performance in adverse weather. However, millimetre-wave technology allows the radar to be deployed in adverse weather. The standalone usage of radar to build the perception system is counterproductive in fully representing the environment due to the sparse nature of data. To rectify this, sensor fusion strategies are being adopted, and we proposed a two-branched framework to fuse the radar and Lidar data for object detection. The primary branch extracts the radar's features, whereas the auxiliary branch extracts the Lidar's features and combines them with radar features by employing additive attention. Later, the embedded features are passed to the novel parallel forked structure (PFS) to handle
% the scale variations. Finally, a region proposal head is adopted for object detection. The proposed method efficacy is evaluated on the Radiate dataset using COCO metrics, outperforming the state-of-the-art methods by a margin of $1.89\%$ and $2.61\%$ in good and adverse weather
% conditions, respectively. This shows that  
% radar-Lidar fusion contributes to accurate object detection and localization in difficult weather conditions.

Object detection is a core component of perception systems, providing the ego vehicle with information about its surroundings to ensure safe route planning. While cameras and Lidar have significantly advanced perception systems, their performance can be limited in adverse weather conditions. In contrast, millimeter-wave technology enables radars to function effectively in such conditions. However, relying solely on radar for building a perception system doesn't fully capture the environment due to the data's sparse nature. To address this, sensor fusion strategies have been introduced. We propose a dual-branch framework to integrate radar and Lidar data for enhanced object detection. The primary branch focuses on extracting radar features, while the auxiliary branch extracts Lidar features. These are then combined using additive attention. Subsequently, the integrated features are processed through a novel Parallel Forked Structure (PFS) to manage scale variations. A region proposal head is then utilized for object detection. We evaluated the effectiveness of our proposed method on the Radiate dataset using COCO metrics. The results show that it surpasses state-of-the-art methods by $1.89\%$ and $2.61\%$ in favorable and adverse weather conditions, respectively. This underscores the value of radar-Lidar fusion in achieving precise object detection and localization, especially in challenging weather conditions.
\end{abstract}

\begin{figure*}[t]
      \centering
      \includegraphics[width=15cm, keepaspectratio]{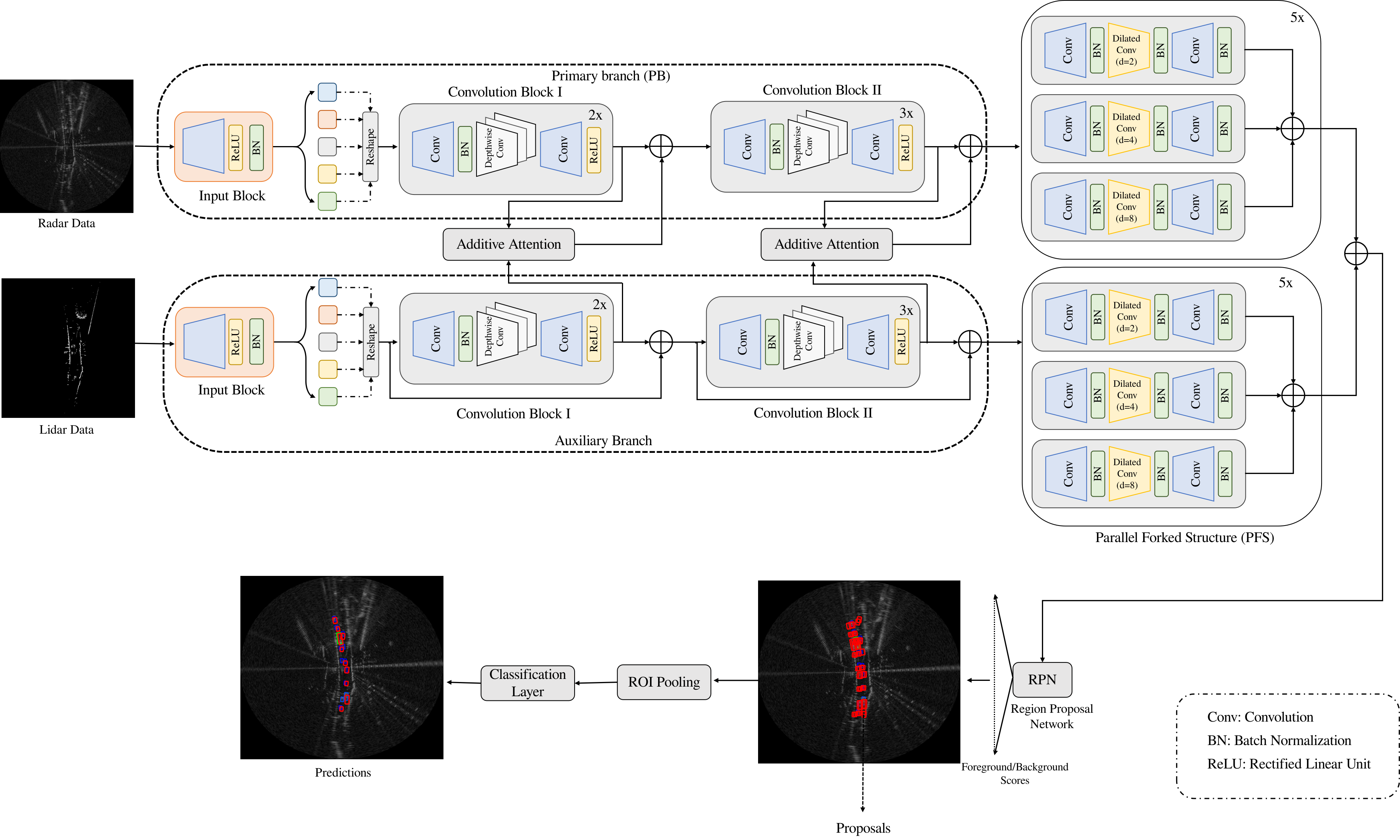}
      \caption{The proposed method for radar-based object detection by fusing the radar and Lidar data. The architecture is composed of primary (PB) and auxiliary branch having the same configurations. Both the branches include input block, convolution block I and II for the features extraction by doing the depth-wise convolution, followed by parallel forked structure (PFS). The data is fused between the two branches using additive attention. Finally, the embedded features from the PFS are combined and used by the region-proposal detection for the object detection.}
      \label{framework}
\end{figure*}
\section{Introduction}
% The perception of the environment plays a critical role in designing safe and reliable decision-making algorithms for the autonomous vehicle \cite{wen2022deep}. Nowadays, autonomous vehicles are equipped with multiple state-of-the-art sensor modalities, for instance, Lidars, cameras and radars \cite{azam2020system} \cite{velasco2020autonomous}. Utilizing these multiple sensor modalities provides accuracy and robustness to perceive the environment yet introduces design challenges to developing unified perception systems for autonomous vehicles. In addition, most of these sensors have limitations in operability \cite{jiao2023msmdfusion}. To mitigate these issues, fusing the multi-modal data is one possible solution in this direction that not only gives us an enriching representation of the environment but also overcomes the individual sensor's limitations \cite{chen2017multi} \cite{kuang2020multi} \cite{zhang2023perception}.

Environmental perception is crucial for crafting dependable decision-making algorithms in autonomous vehicles \cite{wen2022deep}. Modern autonomous vehicles are outfitted with a variety of advanced sensors, including Lidars, cameras, and radars \cite{azam2020system} \cite{velasco2020autonomous}. While leveraging these diverse sensor modalities enhances accuracy and robustness in environmental perception, it also poses challenges in creating cohesive perception systems for such vehicles. Moreover, many of these sensors face operational constraints \cite{jiao2023msmdfusion}. Fusing data from multiple modalities offers a potential solution, providing a richer environmental representation and addressing the individual limitations of each sensor \cite{chen2017multi} \cite{kuang2020multi} \cite{zhang2023perception}.
\par
In the literature, most existing sensor fusion approaches discuss the fusion of Lidar and camera. Although these two sensors give rich information about the environment, yet there are some limitations in using them. For instance, the point cloud from the Lidar gives dense $3$D measurement at a close range but becomes sparse at long ranges, thus limiting its ability to detect objects at a far length \cite{zhang2022not}. Similarly, cameras provide a rich spatial representation of the environment but do not provide depth information compared to Lidar data. However, the fusion of Lidar and camera complement each other and give promising results in 3D object detection for autonomous driving, both of these sensors are prone to adverse weather conditions, which can significantly reduce their performance \cite{chen2017multi}  \cite{kuang2020multi}. %Conversely, radars use electromagnetic waves to determine the object's range, velocity and angle. Primarily operating at millimetre wavelength, it can penetrate or diffract around the tiny particles in adverse weather conditions such as fog, snow, rain and dust, thus can be utilized for the long-range perception of the environment \cite{qian2021robust} \cite{li2022exploiting} \cite{li2022modality}.
Conversely, because radars operate at millimetre wavelength, they can penetrate or diffract around the tiny particles in adverse weather conditions such as fog, snow, rain and dust, thus can be utilized for the long-range perception of the environment
\cite{qian2021robust} \cite{li2022exploiting} \cite{li2022modality}.
\par
Recently, some research efforts have been made to use radar data as a single modality for object detection. Some common approaches that utilize radar data either represent the radar data in image format or as point cloud data. For instance, \cite{azam2021channel} have adopted the radar image representation and designed a channel boosting approach in an ensemble feature extractor network for feature representation. Later, they utilized the transformer network to learn the contextual representation between the embedded features and used them for object detection. Similarly, \cite{li2022exploiting} have transformed the radar data into an ego-centric bird-eye-view radar image to exploit the temporal relations and used them for object recognition. However, majority of  works have utilized  radar data for object detection only as a point cloud representation. That said, these approaches provide promising improvement in radar-based object detection, yet  usage of standalone radar for the environment perception is limited due to its low spatial resolution. To address this, radar data can be fused with Lidar and camera to complement each other \cite{chu2023mt}. The camera-radar fusion improves the spatial resolution of the perception system but results in the sparse representation of the environment.\cite {nabati2021centerfusion}. Some works exploit this and fuse the radar and Lidar data for object detection \cite{qian2021robust}. However, sensor fusion between radar and Lidar provides a better representation of the environment, but how to fuse the data between these two modalities is an open research problem.
\par
% The neural network's hierarchical feature representation permits the design of multiple strategies for fusing the sensor data \cite{feng2020deep}. The most common approaches that have been adopted fall in early, late or mid-fusion schemes. In the early fusion, the sensor data are fused at the data level before extracting the features. This scheme helps the feature extractor network learn the sensor modalities' joint representation. However, the early fusion strategies are sensitive to spatial and temporal misalignment of the data \cite{feng2020deep}. On the other hand, late fusion approaches fuse the sensor's data at a decision level and, thus, therefore provide more room for introducing new sensor modalities to the feature extraction network. The common problem in the late fusion scheme is that the full potential of sensor modalities is not exploited because the joint representation between the sensor's data is not appropriately learnt \cite{feng2020deep}. However, the mid-fusion compromises early and late fusion, enabling the network to learn the joint representation between the sensor modalities \cite{feng2020deep}. 
% A comprehensive analysis is done in related work on data fusion strategies, and to this end, in this work, we have adopted the mid-fusion approach and designed a novel framework for fusing radar and Lidar data for object detection.
In the related work, a comprehensive analysis has been conducted on various data fusion strategies. Based on this analysis, the mid-fusion approach has been adopted in this particular work.  In this work, a novel framework has been specifically designed for fusing radar and Lidar data for the task of object detection.The radar data allows to perceive the environment in adverse weather conditions, while Lidar data provides a dense point cloud of the environment, thus providing a unified framework for object detection.  The proposed methodology, depicted in Fig. \ref{framework}, consists of a pair of branched networks. The primary branch (PB) is responsible for extracting features from the radar data, while the secondary branch leverages Lidar data to provide supplementary information in the form of extracted features. The architectural configuration of both branches remains consistent: an initial input representation block is employed to partition the input data into patches, succeeded by multiple convolutional blocks and a parallel forked structure (PFS). Within the convolutional blocks, we employ depth-wise convolution to extract  features from each channel. This strategy is chosen due to the inherent data sparsity \cite{wang2022can}. The features extracted from these two branches are fused at an intermediate stage by using  additive attention. Subsequently,  the embedded features from both branches are concatenated and passed to a region-based proposal detection head for generating detection results. Similar to work, \cite{qian2021robust} have also used radar-Lidar fusion for object detection, yet the main differences to our work are that they have opted to use late fusion and only consider one specific adverse weather condition: foggy weather. 
The main contributions of our work are:
\begin{enumerate}
\item We designed a novel framework for radar-based object detection by fusing the Lidar information at an intermediate level.
\item We employed additive attention to fuse the data between radar and Lidar, where the Lidar data provide auxiliary information that helps in radar-based object detection. 
\item The proposed network introduces the parallel fork structure to effectively handle the scale variations by changing the receptive field size using the dilated convolution. 
\end{enumerate}
The rest of the paper is organized as follows: Section II covers the literature review. Section III presents the proposed  framework, and Section IV presents experimentation, analysis and results. Finally, Section V concludes the paper with possible directions for future work. 

\section{Related Work}
%In literature, camera and Lidar, despite being widely used to perceive the environment in autonomous driving, however many works have focused on using radar as a single modality for object detection. 
\subsection{Radar Object Detection}
Even though cameras and Lidars are fundamental for perception in autonomous driving in recent years we can observe an influx of works focusing on the radar as a single modality for object detection. Earlier work is focused on classical radar processing techniques involved in detecting the peaks in the radar sensor data. The resulting peaks are clustered and tracked over time and then classified by a single-layer neural network \cite{dickmann2015making}. Similarly, \cite{lombacher2017object} have accumulated the radar peaks in a 2D grid and then adopted a random forest and convolution neural network to classify them in an ensemble network. Furthermore, deep learning approaches have also been studied for radar-based object detection. \cite{major2019vehicle} have proposed a deep learning-based approach for vehicle detection using range-azimuth-doppler measurements. \cite{azam2021channel} have adopted the channel boosting method in an ensemble learning framework for extracting the features from radar data and then employed the transformer network for object detection. Similarly, \cite{li2022exploiting} have first transformed the radar data into ego-centric radar image frames and then exploited the temporal relations for object detection. All of the aforementioned object detection algorithms have used only radar data, but our work focuses on fusion of Lidar and radar data. 
\subsection{Fusion algorithms}
% The neural network's hierarchical feature representation permits the design of multiple strategies for fusing the sensor data \cite{feng2020deep}. The most common approaches that have been adopted fall in early, late or mid-fusion schemes. In the early fusion, the sensor data are fused at the data level before extracting the features. This scheme helps the feature extractor network learn the sensor modalities' joint representation. However, the early fusion strategies are sensitive to spatial and temporal misalignment of the data \cite{feng2020deep}. On the other hand, late fusion approaches fuse the sensor's data at a decision level and, thus, therefore provide more room for introducing new sensor modalities to the feature extraction network. The common problem in the late fusion scheme is that the full potential of sensor modalities is not exploited because the joint representation between the sensor's data is not appropriately learnt \cite{feng2020deep}. However, the mid-fusion compromises early and late fusion, enabling the network to learn the joint representation between the sensor modalities \cite{feng2020deep}. 
Neural networks allow for diverse sensor data fusion strategies due to their hierarchical feature representation \cite{feng2020deep}. Commonly, these strategies include early, mid, and late fusion. Early fusion combines data before feature extraction but can face misalignment issues. Late fusion works at the decision level, but may not capture the full potential of joint sensor data representations. Mid-fusion strikes a balance between the two, effectively learning joint representations of sensor modalities \cite{feng2020deep}.
Some works have fused the radar data with Lidar and camera to complement the sensor's individual limitations. \cite{nabati2021centerfusion} have fused the radar and camera by first finding the centre points in image data using the centre point detection network. Then, it solves the data association problem between radar detections and the object's centre using a frustum-based method. Later, the associated radar features are used for radar object detection. Another work, RODNet \cite{wang2021rodnet} is a cross-supervised camera-radar fusion framework for radar object detection. It uses the teacher-student network, where the teacher network fuses both the radar images and RGB images to obtain the object classes and location in the radar images. Finally, the teacher network supervises the student network that only uses radar images for object detection. MT-DETR \cite{chu2023mt} has proposed a multi-modal fusion network by introducing the residual and confidence fusion modules to fuse the radar, camera and Lidar. In addition, a residual enhancement module is designed to enhance each unimodal branch's feature extraction. Similarly, \cite{nobis2021radar} have also fused the Lidar, camera and radar for object detection. MVDNet \cite{qian2021robust} has fused the Lidar and radar by first generating the proposal from two sensor data and then adopting the region-wise features to improve the object detections. Similarly, ST-MVDNet \cite{li2022modality} is built on top of MVDNet \cite{qian2021robust} by introducing the teacher-student network for radar-based object detection. In their framework, the teacher model generates the predictions to train the student network using a consistency constraint while the student network passes the parameters it learns to the teacher network via an exponential moving average. 
\par
MVDNet \cite{qian2021robust} and ST-MVDNet \cite{li2022modality} are more similar to our work employing radar and Lidar fusion for object detection. However, both of these methods only consider foggy weather conditions. In addition, MVDNet \cite{qian2021robust} adopts a late fusion approach that limits learning the joint distribution of sensor modalities in contrast to our mid-fusion approach. However, ST-MVDNet \cite{li2022modality} uses the knowledge distillation approach, which has a problem of large discrepancy of predictive distribution between teacher and student networks. To address these issues, we designed a novel fusion network that uses additive attention for sensor fusion and depth-wise convolution to extract useful features from the sparse data of radar and Lidar, respectively. In addition, we introduce a parallel fork structure to handle the scale variations that solve the predictive discrepancy between the learned features.

% In this section, we explain in detail the proposed framework for object detection using Lidar and Radar data as illustrated by Fig. \ref{}. The proposed framework leveraged additive attention in the backbone design to learn the representation from input sources and employed a parallel forked structure( PFS), where each unit has the same transformation parameters but varying receptive fields. The PFS renders scale-specific feature maps with a consistent representation of features. 

\section{Proposed method}
In this section, we explain in detail the proposed multi-modal fusion framework for object detection using Lidar and radar data, as illustrated in Fig. \ref{framework}. 
% The framework consists of a branched network; the PB learns feature representation from radar data and consists of four blocks: the input block, convolution blocks I \& II and the parallel fork structure (PFS) block. The proposed method adopts an auxiliary branch following the same architectural design as the PB, used for feature extraction from the Lidar data. Later, the auxiliary branch features are fused with radar features using additive attention. The embedded features from both primary and auxiliary branches are fused and passed to the region-proposal-based detection head for object detection.
% The second branch extracts feature from lidar data and have a similar configuration to the radar branch.
% However, we leverage additive attention to fuse the Lidar information from the auxiliary branch. The output from branches is finally stacked and given to the detection head to perform object detection.
\subsection{Input  Block}
Effective feature representation is crucial for deep learning-based object detection. The key idea is to explore the occurrence of features from multi-modal data by designing the neural network backbone for learning robust feature representations.  In our proposed method settings, the inputs are represented in an image format. The radar image is acquired from the sensor, capturing information about the surrounding objects. Additionally, the Lidar data is preprocessed to obtain an image representation. The Lidar point-cloud data is preprocessed to a bird-eye-view (BEV) representation to make it more applicable to be used by $2$D convolution neural networks in our proposed method. The majority of the works in object detection use a standard convolutional stem with a large kernel size as the input layer, which down-samples the input to acquire a proper feature map \cite{ren2015faster} \cite{lin2017focal}. However, we have opted for a patch-based representation of the input data, where the input data is split into fixed-size patches and obtained linear embeddings, which preserve locality and reduces computations complexity\cite{dosovitskiy2020image}\cite{liu2022convnet}. The input image pair is given by ${P=(I_i, J_i)}$, where ${I \in \mathbb{R}^{H\times W\times C}}$ represent the radar range-azimuth images in cartesian coordinate system and ${J \in \mathbb{R}^{H\times W\times C}}$ shows the Lidar BEV image. 
The image ${I}$ is partitioned into ${p \times p}$ non-overlapping patches ($x_p$) given by ${x_p \in \mathbb{R}^{N \times(P^2.C)}}$, where $(H, W)$ is the height and width of the original image, $C$ represent the input channels, $p$ is the patch-size and $N=HW/p^2$ are the number of patches. The Eq.\ref{eq1} gives the linear embedding, which is further given to the convolution blocks, where $\zeta$ denotes nonlinear activation function.
\begin{equation}
    x_0=BN(\zeta \{Conv2D(x,stride=P,Kernel=P)\})
    \label{eq1}
\end{equation}

% \usepackage{booktabs}
% \usepackage{graphicx}
\begin{table}[b]
\caption{The layer detail of the architecture for the proposed framework.}
\centering
\label{layer }
\resizebox{7cm}{!}{%
\begin{tabular}{@{}lcc@{}}
\toprule
Blocks                                          & \multicolumn{1}{c}{Input/Output channel size} & \multicolumn{1}{c}{Layers}                                                                                                                                                                                                                                     \\ \midrule
\multicolumn{1}{l|}{input block}               & \multicolumn{1}{c|}{$3/64$}                   & \multicolumn{1}{c}{\begin{tabular}[c]{@{}c@{}}$\begin{bmatrix}\\ 16 \times 16 , 64\\ RELU \\   BN \\ \end{bmatrix}$\end{tabular}}                                                                                                                          \\ \midrule
\multicolumn{1}{l|}{convolution block I}       & \multicolumn{1}{c|}{$64/256$}                 & \multicolumn{1}{c}{\begin{tabular}[c]{@{}c@{}}$\begin{bmatrix}\\ 1 \times 1,256 \\ BN \\ 11 \times 11,256 \\ RELU\\ \end{bmatrix} \times 3$\end{tabular}}                                                                                                 \\ \midrule
\multicolumn{1}{l|}{Convolution block II}      & \multicolumn{1}{c|}{$256/512$}                & \multicolumn{1}{c}{\begin{tabular}[c]{@{}c@{}}$ \begin{bmatrix}\\ 1 \times 1, 512 \\ BN \\ 11 \times 11, 512 \\ 1 \times 1, 512  \\ RELU\\ \end {bmatrix} \times 4$\end{tabular}}                                                                     \\ \midrule
\multicolumn{1}{l|}{Parallel Forked Structure} & \multicolumn{1}{c|}{$512/1024$}               & \multicolumn{1}{c}{\begin{tabular}[c]{@{}c@{}}$ \begin{bmatrix}\\ 1 \times 1,512 \\ BN \\ \{11 \times 11,d=2,512 \} \{11 \times 11,d=4,512 \}\\ \{11 \times 11,d=8,512 \} \\ BN \\ 1 \times 1,1024 \\ RELU\\ \end {bmatrix} \times 5$\end{tabular}} \\ \bottomrule
\end{tabular}%
}
\end{table}
\subsection{Convolution Blocks I \& II }
Multi-modal object detection encounters challenges in answering the central question, which is the best way to fuse information from different modalities. Although, in literature, early, mid and late fusion techniques have been explored; it has been shown, in studies \cite{chitta2022transfuser} \cite{meng2020survey}, mid-fusion techniques are most efficient in determining characteristics that co-occur and assist in acquiring rich features for object detection. Therefore, we also follow mid-fusion architecture, where we use two feature extractor branches, one for each modality, consisting of convolutional blocks I \& II, and then leverage additive attention to fuse the features as illustrated in Fig. \ref{framework}.
Table \ref{layer } gives the details of each convolution block, which consists of point-wise convolutions, BatchNorm, depth-wise convolution and nonlinear activation. The depth-wise convolution is adopted to learn the spatial locality of features from the sparse data, which assists in learning better local and global feature representation from each data channel and using large kernel size provides robustness on out-of-distribution samples\cite{liu2022convnet}\cite{xie2017aggregated}. 
 The output from the convolutional block of the radar ($x_r$) and Lidar ($x_l$) branches is fused using additive attention.\cite{bahdanau2014neural}. The output from additive attention is added to the PB since we are using ground-truth labels in the radar reference frame. Using additive attention enhances the feature representation and localization, reducing the feature response in unrelated background parts without the need to extract the area of interest. 
 \par
\begin{figure}[]
      \centering
      \includegraphics[width=6cm]{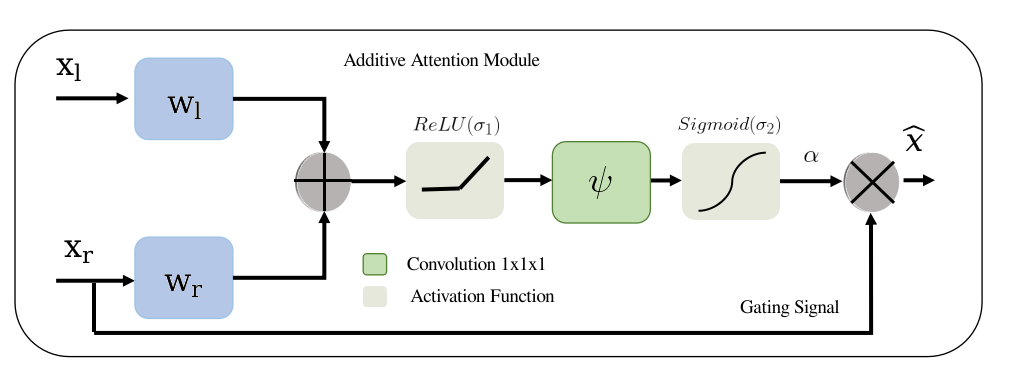}
      \caption{The additive attention working operation to fuse radar and Lidar feature maps.}
      \label{addtive_attention}
\end{figure}

 Fig. \ref{addtive_attention} shows the  operation of additive attention. The fusion of features through attention provides flexibility regarding what to focus on from a regional basis when fusing the information with the gating signal. The additive attention coefficient $\alpha \in [0,1]$ distinguishes prominent feature map regions and prunes features for task-specific activations \cite{shen2018disan}. The output of the additive attention module is the element-wise multiplication of the fused feature map and input feature map described in Eq. \ref{hat}. 
\begin{equation}
       \hat{x} =x_r.\alpha
       \label{hat}
\end{equation}
The $\alpha$ is given by Eq.\ref{alpha}.
\begin{equation}
\label{alpha}
\begin{aligned}
& x_{att}=\psi^T (\sigma (W^T_{r}{x}_{r}+W^T_{l}{x}_{l}))+b_\psi, \\
& \alpha =\sigma _2(x_{att}({x}_{r},{x}_{l,i};\Theta_{att} )). 
\end{aligned}
\end{equation}
where $\sigma _2({x})=\frac{1}{1+exp(-x)}$ is the sigmoid activation function. $\Theta_{att} $ characterizes the set of learning parameters including $W_{r}$, $W_{l}$, and $\psi$, and the bias term, $b_g$. These parameters are trained with standard backpropagation.
\subsection{Parallel Forked Structure (PFS) Block }
This section explains the parallel fork structure, which handles scale variation for object detection by varying the receptive field while having the same transformation parameters. As a result, the PFS renders scale-specific feature maps with a consistent representation of features. Contrary to our work, the most popular techniques to handle scale variation in object detection are image pyramids \cite{adelson1984pyramid} and feature pyramids. The image pyramid introduces computational inefficiency but does provide representational power to deal with objects of all scales equally. However, the feature pyramid sacrifices the feature consistency across different scales leading to a higher risk of overfitting \cite{li2018detnet}.
\par 
The PFS block consists of three parallel forked branches, each with a different receptive field, implemented using dilated convolution \cite{yu2015multi}. Dilation convolutions enhance feature expression and reduce computation by selectively spacing convolutional kernel elements, enabling a wider receptive field with fewer parameters. This helps to  effectively captures intricate patterns and contextual information, showcasing the superior efficiency and effectiveness of this convolutional technique. The dilated convolution for one-dimensional input data is given by Eq. \ref{dia}, where input  $x_i$, output $y_i$, kernel filter $w_k$ of the length of k and $d$ is the dilated parameter that corresponds to stride through the input feature. 
\begin{equation}
y_i=\sum_{k=1}^{K} x_{i+d.k}.w_k,
\label{dia}
\end{equation}
The standard convolution is dilated convolution with $d=1$. $i$ variable shows the location on output feature $y_i$, when kernel filter $w_k$ is applied on input feature $x_i$. $k$ denotes the indices of the dilated convolution kernel. The dilated convolution with rate $d$ introduces $d-1$ zeroes in the consecutive filter values, increasing the kernel size of the $k \times k $ filter to $k_d=k+(k-1)(d-1)$ with no increase in computation power and the same number of parameters. Therefore, it provides an effective method for handling scale variation and finding the best concession between context assimilation and finding the region of interest. 

\subsection{Detection Head}
The outputs from the multi-modal branches are fused and passed through $1 \times 1$ convolution to learn the combined features for the downstream object detection task. The output of this convolution is passed to the region proposal network \cite{ren2015faster}, which predicts the bounding box and class for each object. Eq. \ref{loss} gives the region proposal network loss. 
\begin{equation}
\label{loss}
    L=\frac{1}{N_{cls}}\sum_{j}L_{cls}(p_j,p_j^*)+\lambda\frac{1}{N_{reg}}\sum_{j}p_j^*(b_j,b_j^*)
\end{equation}
where, $j$ denotes the anchor index, $p_j$ is predicted probaility of objectness in an anchor, $p_j^*$ is the ground truth. $b_i$ denotes the bounding box coordinates and $b_i^*$ represent the ground truth values. The classification loss $L_{class}$ is log loss, and regression loss $L_{reg}$ is defined by smooth $L1$ loss. $N_{cls}$ and $N_{reg}$ are the normalization term and $\lambda$ is weighted term which is set to $\lambda =
10$.

\section{Experimentation and Results}
\subsection{Dataset}

In this study, we have used the RADIATE (RAdar Dataset In Adverse weaThEr)  dataset \cite{sheeny2021radiate}. The RADIATE dataset is collected in adverse weather conditions and covers numerous traffic scenarios, facilitating object detection research in dynamic environments under adverse weather. The perception sensor data is collected using the Navtech CTS350-X radar and 32-channel Velodyne Lidar. The radar generates $360^\circ$ range-azimuth images with a $1152 \times 1152$ resolution. However, Lidar renders a point cloud that has $360^\circ$ coverage. The sensors are calibrated by finding extrinsic parameters for Lidar and radar. The radar is used as the reference frame to calculate the parameters. 
The Lidar point-cloud data is pre-processed to obtain bird-eye-view (BEV) two-dimensional images by using binary occupancy grid mapping. For this, the BEV grid of $100m \times 100m$ grid is considered, where the block size of the grid is $0.174m \times 0.174m$, which results in a BEV image of $576 \times 576$ pixels. To further enhance the BEV images without data loss, the resulting BEV image is up-sampled to give the resolution of $1152 \times 1152$ pixels having $3$ channels, representing occupancy, height, and intensity information obtained from the Lidar point cloud. We opted for BEV representation for input data compared to 3D voxelization of point cloud data, as it is more computationally effective and preserves the metric space, allowing the framework to explore features about the size and shape of objects. 

The dataset contains data from different weather conditions, for instance, fog, rain, snow, night, sunny and cloudy. Moreover, it has various driving scenarios like urban, suburban and motorway. The annotated data consists of 40K images with an estimated 200K labelled vehicles. The vehicles are annotated using a bounding box defined by $(x, y, w, h, angle)$. Table \ref{table3} shows the training and test split of the dataset.

\begin{table}[]
\centering
\caption{The RADIATE dataset partition topology for training and testing the proposed approach in adverse weather conditions }
\label{table3}
\begin{tabular}{@{}l|c|c@{}}
\toprule
 & \multicolumn{1}{l|}{No. of Images} & \multicolumn{1}{l}{No. of vehicles} \\ \midrule
Training data \textit{Good Weather}         & 23091 & 106931 \\
Training data  \textit{Good \& Bad Weather} & 9760  & 39647  \\
Testing  data                             & 11289 & 147005 \\ \bottomrule
\end{tabular}
\end{table}
\subsection{Experimentation Details}
The proposed framework is trained on two RTX3090 GPUs using the PyTorch library. The input training data is augmented, scaled, randomly flipped and randomly cropped to provide more generalized datasets. The network is trained for 100K iterations using a batch size of 4. A stochastic gradient descent (SGD) optimizer is used with a learning rate of 0.02, and a linear scheduler is incorporated to decrease the learning rate by a factor of 0.1 after 1K iterations. 
The input patch-embedding size is 16, and the depthwise convolution kernel size is 11. The parallel fork structure consists of 3 branches with 2, 4 and 8 dilation rates. The region proposal network obtained about 12000 proposals from the parallel forked structure, and 128 ROI were sampled for training. The standard COCO evaluation metric of Average precision (AP) is used for evaluation \cite{lin2014microsoft}.The higher value of AP score corresponds to
better object detection model. 
% Learning how to represent features effectively is crucial in deep-learning-based object detection. The key idea is to explore the co-occurrence of information from multi-modal data by designing the neural network backbone for learning robust feature representations. The input images use patch-based representation, followed by the convolution blocks employing depthwise convolution. However, additive attention is used to fuse the features after each convolution block from two branches of multi-modal data. The last convolution block introduces a parallel fork structure, which learns scale-aware feature map.

% \usepackage{booktabs}
% \usepackage{multirow}
% \usepackage{graphicx}
\begin{table}[]
\caption{Quantitative analysis of proposed method using only radar and fusion of radar and Lidar is performed. The analysis illustrates the mAP scores of proposed method and state-of-the-art methods in both good, and adverse weather conditions.}
\centering
\label{results}
\resizebox{8cm}{!}{%
\begin{tabular}{@{}ll|cl|cl|@{}}
\cmidrule(l){3-6}
 &
   &
  \multicolumn{2}{c|}{Split: train good weather} &
  \multicolumn{2}{l|}{Split: train good and bad weather} \\ \cmidrule(l){2-6} 
\multicolumn{1}{l|}{} &
  Methods &
  \multicolumn{2}{c|}{mAP@0.5 scores} &
  \multicolumn{2}{c|}{mAP@0.5 scores} \\ \midrule
\multicolumn{1}{l|}{\multirow{13}{*}{\centering{\rot{Radar}}}} &
  RetinaNet-OBB-ResNet18 \cite{li2022exploiting} &
  \multicolumn{2}{c|}{37.83} &
  \multicolumn{2}{c|}{31.57} \\ \cmidrule(l){2-6} 
\multicolumn{1}{l|}{} &
  RetinaNet-OBB-ResNet34 \cite{li2022exploiting} &
  \multicolumn{2}{c|}{35.61} &
  \multicolumn{2}{c|}{31.10} \\ \cmidrule(l){2-6} 
\multicolumn{1}{l|}{} &
  RetinaNet-OBB-ResNet34-T \cite{li2022exploiting} &
  \multicolumn{2}{c|}{37.30} &
  \multicolumn{2}{c|}{24.50} \\ \cmidrule(l){2-6} 
\multicolumn{1}{l|}{} &
  CenterPoint-OBB-EfficientNetB4 \cite{li2022exploiting} &
  \multicolumn{2}{c|}{51.43} &
  \multicolumn{2}{c|}{42.37} \\ \cmidrule(l){2-6} 
\multicolumn{1}{l|}{} &
  CenterPoint-OBB-ResNet18 \cite{li2022exploiting} &
  \multicolumn{2}{c|}{49.41} &
  \multicolumn{2}{c|}{44.48} \\ \cmidrule(l){2-6} 
\multicolumn{1}{l|}{} &
  CenterPoint-OBB-ResNet34 \cite{li2022exploiting} &
  \multicolumn{2}{c|}{50.17} &
  \multicolumn{2}{c|}{42.81} \\ \cmidrule(l){2-6} 
\multicolumn{1}{l|}{} &
  BBAVectors-ResNet18 \cite{li2022exploiting} &
  \multicolumn{2}{c|}{50.53} &
  \multicolumn{2}{c|}{45.43} \\ \cmidrule(l){2-6} 
\multicolumn{1}{l|}{} &
  BBAVectors-ResNet34 \cite{li2022exploiting} &
  \multicolumn{2}{c|}{51.26} &
  \multicolumn{2}{c|}{44.61} \\ \cmidrule(l){2-6} 
\multicolumn{1}{l|}{} &
  TRL-EfficientNetB4 \cite{li2022exploiting} &
  \multicolumn{2}{c|}{50.98} &
  \multicolumn{2}{c|}{43.05} \\ \cmidrule(l){2-6}
\multicolumn{1}{l|}{} &
  TRL-ResNet18 \cite{li2022exploiting} &
  \multicolumn{2}{c|}{53.11} &
  \multicolumn{2}{c|}{46.42} \\ \cmidrule(l){2-6} 
\multicolumn{1}{l|}{} &
  TRL-ResNet34 \cite{li2022exploiting} &
  \multicolumn{2}{c|}{54.00} &
  \multicolumn{2}{c|}{43.98} \\ \cmidrule(l){2-6}
\multicolumn{1}{l|}{} &
  Radiate-ResNet50 \cite{sheeny2021radiate} &
  \multicolumn{2}{c|}{45.31} &
  \multicolumn{2}{c|}{45.77} \\ \cmidrule(l){2-6} 
\multicolumn{1}{l|}{} &
  Radiate-ResNet101 \cite{sheeny2021radiate} &
  \multicolumn{2}{c|}{45.84} &
  \multicolumn{2}{c|}{46.55} \\ \cmidrule(l){2-6}  
\multicolumn{1}{l|}{} &
  Channel-Boosted-ResNet50 \cite{azam2021channel} &
  \multicolumn{2}{c|}{54.90} &
  \multicolumn{2}{c|}{55.54} \\ \cmidrule(l){2-6} 
\multicolumn{1}{l|}{} &
  Channel-Boosted-ResNet101 \cite{azam2021channel} &
  \multicolumn{2}{c|}{58.39} &
  \multicolumn{2}{c|}{59.03} \\ \cmidrule(l){2-6} 
\multicolumn{1}{l|}{} &
  Ours (Radar only) &
  \multicolumn{2}{c|}{\textbf{63.45}} &
  \multicolumn{2}{c|}{\textbf{60.95}} \\ \midrule
\multicolumn{1}{l|}{\multirow{4}{*}{\rot{Radar+Lidar}}} &
  MVDNet \cite{qian2021robust} &
  \multicolumn{2}{c|}{60.57} &
  \multicolumn{2}{c|}{59.35} \\ \cmidrule(l){2-6} 
\multicolumn{1}{l|}{} &
  ST-MVDNet \cite{li2022modality} &
  \multicolumn{2}{c|}{65.25} &
  \multicolumn{2}{c|}{62.47} \\ \cmidrule(l){2-6} 
\multicolumn{1}{l|}{} &
  MT-DETR (w/o camera stream) \cite{chu2023mt} &
  \multicolumn{2}{c|}{64.95} &
  \multicolumn{2}{c|}{61.89} \\ \cmidrule(l){2-6} 
\multicolumn{1}{l|}{} &
  Ours (Radar+Lidar) &
  \multicolumn{2}{c|}{\textbf{67.14}} &
  \multicolumn{2}{c|}{\textbf{65.08}} \\ \bottomrule
\end{tabular}%
}
\end{table}

\begin{figure}[t]
      \centering
      \includegraphics[width=7cm]{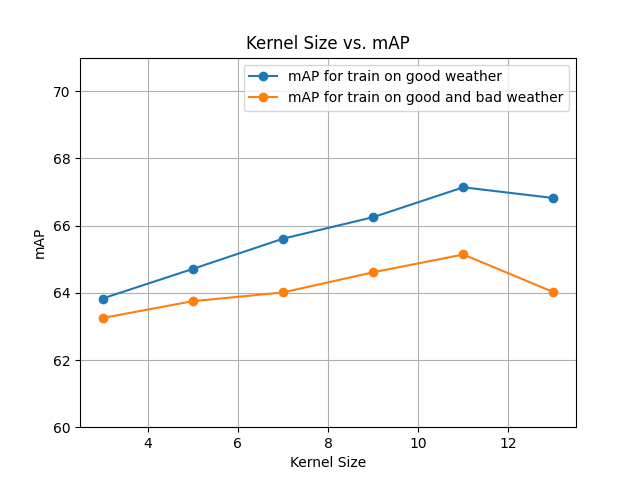}
      \caption{The results on test data for using different kernel sizes in depthwise convolution.}
      \label{kernel}
\end{figure}

\begin{table}[]
\caption{The table shows the result for the early and late fusion of multi-sensor data in comparison to mid-fusion.}
\label{fusion}
\resizebox{8cm}{!}{%
\begin{tabular}{@{}l|c|c@{}}
\toprule
Model                 & split: train good weather & Split: train good and bad weather \\ \midrule
Early Fusion          &  61.59                         &   59.10                                \\ \midrule
Late Fusion           &   63.36                        &    61.47                               \\ \midrule
Mid-fusion (Proposed) & 67.14                     & 65.08                             \\ \bottomrule
\end{tabular}%
}
\end{table}

\begin{figure*}[]
      \centering
      \includegraphics[width=15cm]{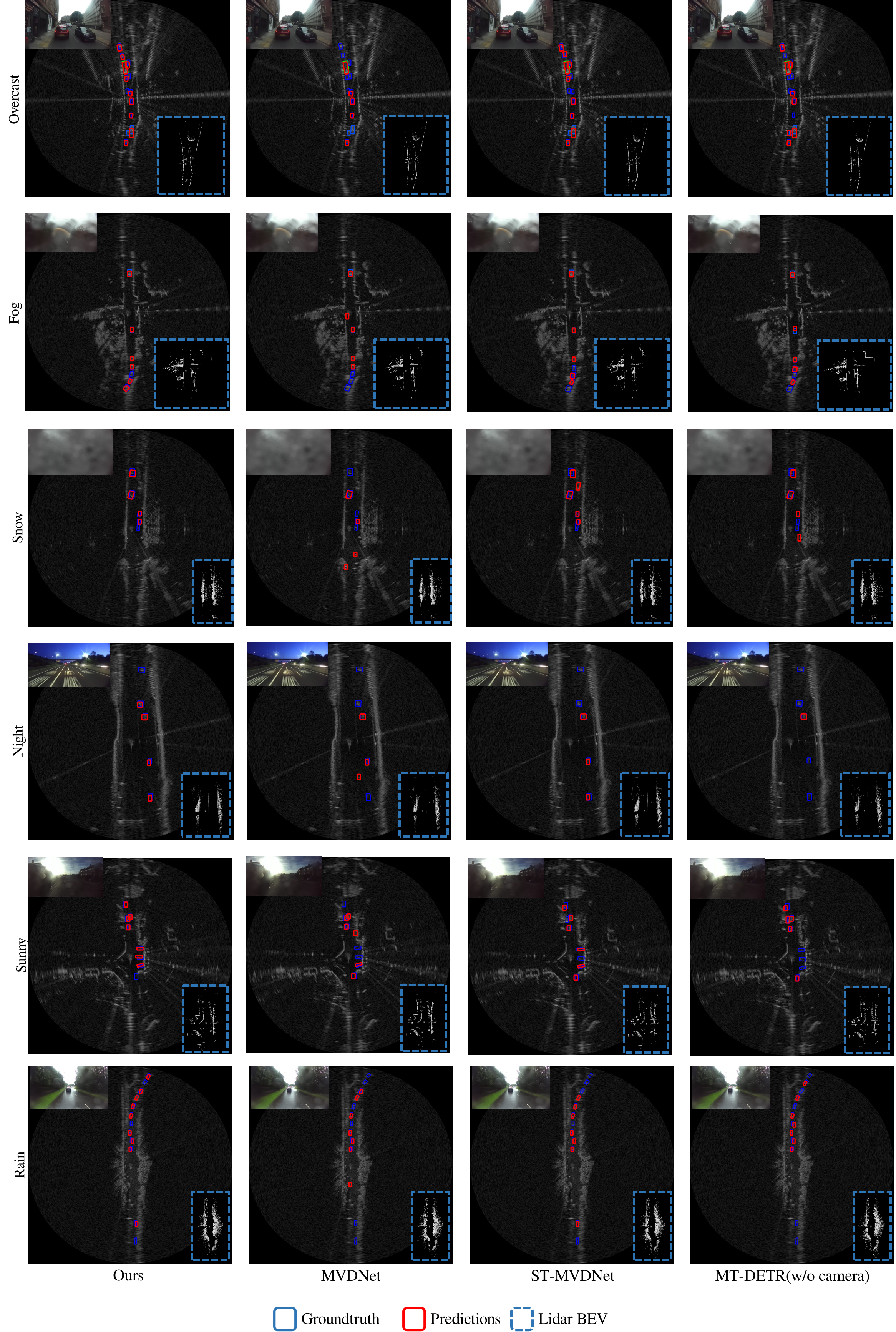}
      \caption{}
      \label{vr}
\end{figure*}

\subsection{Results}

The proposed method is quantitatively and qualitatively evaluated on the Radiate dataset \cite{sheeny2021radiate}, having two different configurations of weather settings: good weather and adverse weather conditions (good and bad weather). In order to quantitatively evaluate the proposed method's efficacy, we first analyzed object detection using radar as a single modality. This quantitative analysis forms a baseline for the proposed method to address how much auxiliary information from the Lidar branch provides robustness in object detection. To this end, the proposed method only with radar input is quantitatively evaluated with the state-of-the-art method that utilizes only the radar data. Table \ref{results} illustrates the quantitative results for the proposed method (only radar) and state-of-the-art methods in both weather conditions settings. The proposed method with only radar has achieved the mAP score of $63.45$ in good weather conditions and $60.95$ in adverse weather conditions, respectively, thus outperforming the best among state-of-the-art method \cite{azam2021channel} by $5.06\%$ in good weather and by a margin of $1.92\%$ in adverse weather conditions respectively. It is to be noted here that we only used the methods that utilize the Radiate dataset for radar-based object detection.
\par
In order to quantitatively evaluate the proposed method in a sensor fusion setting, we employed the state-of-the-art methods developed using other datasets. To the best of our knowledge, no work has used the Radiate dataset \cite{sheeny2021radiate} for fusing radar and Lidar data in sensor fusion settings. To this end, we re-implemented and refactored the codebase to be utilized by the proposed method using the Radiate dataset. The state-of-the-art methods that are used for the quantitative evaluation are MVDNet \cite{qian2021robust}, ST-MVDNet \cite{li2022modality} and MT-DETR (without camera stream) \cite{chu2023mt}. MVDNet \cite{qian2021robust} is a two-stage fusion network where region proposals are generated for both sensor modalities and fused together for object detection. MVDNet \cite{qian2021robust} uses Oxford Radar Robotcar (ORR) dataset \cite{barnes2020oxford} for the evaluation. Similarly, ST-MVDNet \cite{li2022modality} is developed on top of MVDNet \cite{qian2021robust}, which employs the teacher-student network for mutual learning of object detection features and is evaluated on the ORR dataset. MT-DETR \cite{chu2023mt} adopts a hierarchical fusion mechanism by designing a residual fusion module, a confidence fusion module a residual enhancement module to learn the features effectively. MT-DETR \cite{chu2023mt} uses the STF dataset in the experimental analysis. However, the original MT-DETR \cite{chu2023mt} uses the camera, Lidar and radar data in the sensor fusion settings, but in our proposed work, we refactored the original MT-DETR only to accept two sensor streams: radar and Lidar. We opted for the same quantitative evaluation metrics of mAP scores and compared the proposed method with the state-of-the-art method in both good and adverse weather conditions. Table \ref{results} shows the quantitative results where the proposed method has achieved the mAP score of $67.14$ in good weather and $65.08$ mAP score in adverse weather conditions, respectively. The proposed method in sensor fusion settings has outperformed ST-MVDNet by a margin of $1.89 \%$ in good weather conditions and by $2.61 \%$ in adverse weather conditions, respectively. The quantitative evaluation of the proposed method clearly shows the improvement when the Lidar auxiliary information is used for radar-based object detection. The addition of auxiliary information has increased the proposed method mAP score by $3.61 \%$ in good weather conditions and similarly by a margin of $4.13 \%$ mAP score improvement in adverse weather conditions. In order to visualize the radar object detection results, the proposed method and state-of-the-art methods qualitative results are illustrated in Fig.\ref{vr} for both good and adverse weather conditions. 

\par 

Furthermore, we investigate the effect of kernel size on the robustness of the proposed framework. Fig.\ref{kernel} shows the mAP scores at different settings of kernel size. The optimal scores are obtained at $11 \times 11$, which provides a larger receptive field to capture the contextual understanding of the feature maps and robustness to the variation due to noise. 
We examine different fusion approaches for the radar and Lidar fusion; Table \ref{fusion} shows the mAP scores for the early, mid and late fusion schemes. The input images were stacked channel-wise and given to the network in early fusion. However, in late fusion, the feature maps were concatenated after the parallel forked structure, and there was no information exchange between the two branches. The mid-fusion approach outperforms the prior strategies, as it provides the network to learn the joint representation of features map between different sensor modalities and also provides robustness against spatial and temporal misalignment of data.

\section{Conclusion}
This study proposed a multimodal fusion network for object detection in adverse weather conditions using radar and Lidar data. A branched architecture is introduced, which has an input representation block, followed by convolution blocks I \& II and the PFS block. The object detections are generated using a region proposal network. The efficacy of the algorithm is evaluated using the COCO metric. The mean average precision of $67.14$ and $65.08$ is achieved on test data for good and good-bad weather, respectively. 
The results show that the mid-fusion network performs better than the early or late fusion of features. Furthermore, using Lidar data improves the detection accuracy, and a large kernel size is adequate for capturing robust feature representation from images. 
In future work, we aim to extend this work to include images and extensively test the framework on other public datasets.

\section*{Acknowledgment}
This work is supported by the Academy of Finland Flagship program: Finnish Center for Artificial Intelligence (FCAI), and also by Culture, Sports and Tourism Research and Development Program through the Korea Creative Content Agency Grant funded by the Ministry of Culture, Sports and Tourism (R2022060001, “Development of Service Robot and Contents Supporting Children’s Reading Activities Based on Artificial Intelligence”).

\bibliographystyle{IEEEtran}
\bibliography{Radiate-ITSC.bib}

\end{document}